\documentclass{article}
\usepackage{ijcai25}

\usepackage{times}
\usepackage{soul}
\usepackage{url}
\usepackage[hidelinks]{hyperref}
\usepackage[utf8]{inputenc}
\usepackage[small]{caption}
\usepackage{graphicx}
\usepackage{amsmath}
\usepackage{amsthm}
\usepackage{booktabs}
\usepackage{algorithm}
\usepackage{algorithmic}
\usepackage[switch]{lineno}
\usepackage{algorithm}
\usepackage{algorithmic}
\usepackage{xcolor}
\usepackage{booktabs}
\usepackage{subcaption} 
\usepackage{amsmath}
\usepackage{listings}

\title{  Noise-Resistant Label Reconstruction Feature Selection for Partial Multi-Label
Learning }

\author{
Wanfu Gao$^{1,2}$
\and
Hanlin Pan$^{1,2}$\and
Qingqi Han$^{1,2}$\And
Kunpeng Liu$^3$\thanks{corresponding author}\\
\affiliations
$^1$College of Computer Science and Technology, Jilin University, China\\
$^2$Key Laboratory of Symbolic Computation and Knowledge Engineering of Ministry of Education, Jilin University, China\\
$^3$Department of Computer Science, Portland State University, Portland, OR 97201 USA\\
\emails
gaowf@jlu.edu.cn, panhl23@mails.jlu.edu.cn, hanqq22@mails.jlu.edu.cn, kunpeng@pdx.edu
}

\begin{document}

\maketitle

\begin{abstract}
The "Curse of dimensionality" is prevalent across various data patterns, which increases the risk of model overfitting and leads to a decline in model classification performance. However, few studies have focused on this issue in Partial Multi-label Learning (PML), where each sample is associated with a set of candidate labels, at least one of which is correct. Existing PML methods addressing this problem are mainly based on the low-rank assumption. However, low-rank assumption is difficult to be satisfied in practical situations and may lead to loss of high-dimensional information. Furthermore, we find that existing methods have poor ability to identify positive labels, which is important in real-world scenarios. In this paper,  a PML feature selection method  is proposed considering two important characteristics of dataset: label relationship's noise-resistance and label connectivity. Our proposed method utilizes label relationship's noise-resistance to disambiguate labels. Then the learning process is designed through the reformed low-rank assumption. Finally, representative labels are found through label connectivity, and the weight matrix is reconstructed to select features with strong identification ability to these labels. The experimental results on benchmark datasets demonstrate the superiority of the proposed method.
\end{abstract}

\section{Introduction}

PML \cite{xie2018partial} is a recently emerging paradigm of weakly supervised learning aiming to construct a multi-class classifier with uncertain data. Specifically, PML attempts to learn the model from partially labeled samples: a sample is assigned with a candidate label set, and at least one label in the candidate label set is truly related to the sample but the total number of truly related labels is unknown \cite{sun2019partial,yu2018feature}. Compared to multi-label learning \cite{li2023multi,gao2023unified}, PML can better handle situations where labels are missing or ambiguous, which is quite common in real-world scenarios.

Currently, methods for PML can be broadly classified into two categories \cite{xie2021partial,wang2019discriminative,durand2019learning}. The first ones treat all labels in  candidate set as correct annotations, ignore noises in labels and directly utilize models for learning such as ML-KNN \cite{zhang2007ml}. The second ones consider the noises in candidate label set and reduce the influence of noises through matrix decomposition and low-rank confidence matrix approximation \cite{sun2019partial}.
\begin{figure}[h!]
    \centering
    \begin{minipage}{0.7\linewidth}
        \includegraphics[width=\textwidth]{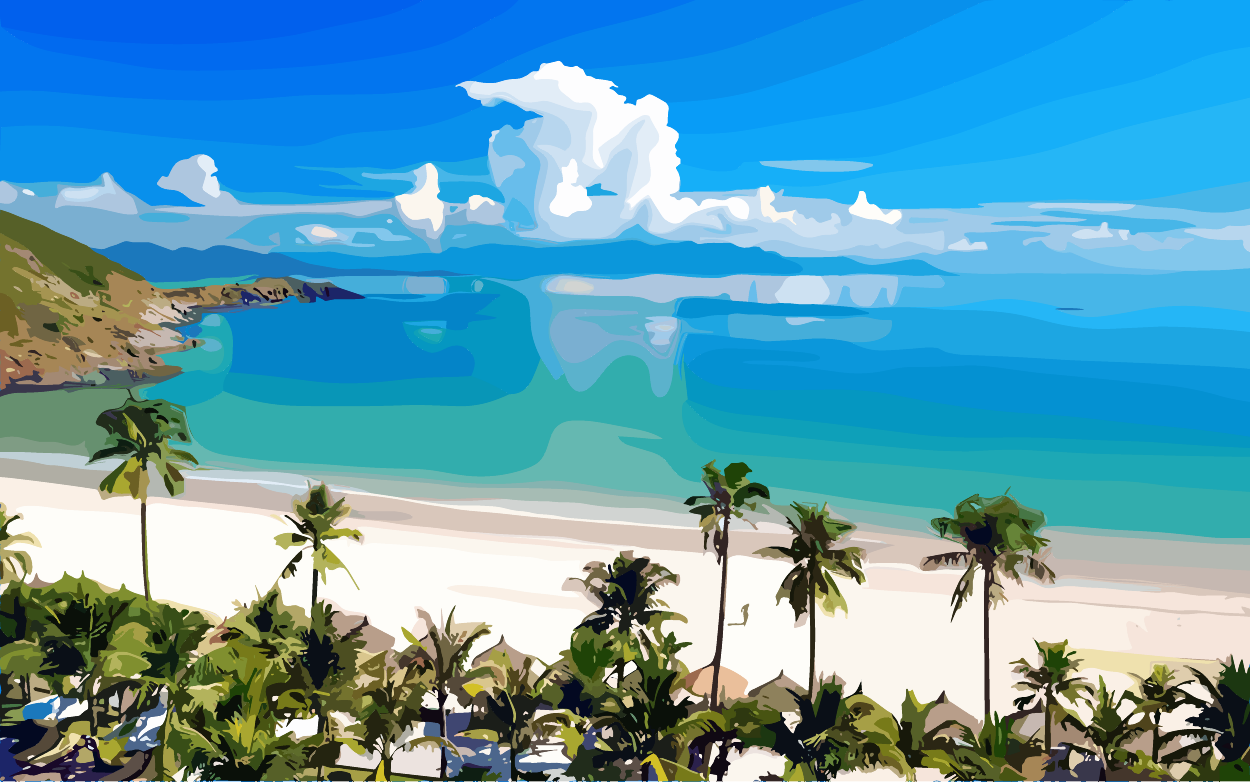}
    \end{minipage}%
    \begin{minipage}{0.3\linewidth}
        \raggedright
        \textbf{Candidate labels}
        \begin{itemize}
            \item \textcolor{red}{cloud}
            \item people
            \item \textcolor{red}{mountain}
            \item \textcolor{red}{tree}
            \item car
            \item \textcolor{red}{sea}
            \item \textcolor{red}{beach}
           
        \end{itemize}
    \end{minipage}
    \caption{An example of PML. Among the candidate set of seven labels, only five of them are valid ones (in red).}
\end{figure}

However, the formers are highly affected by noises, while the latters often rely on the low-rank assumption (i.e. assuming that the low-rank matrix generated after 
dimension reduction is consistent with the original high-order matrix) and reduce the dimension of the feature/label matrix to reduce the influence of noises. While previous studies have shown that the requirements of the low-rank assumption are relatively strict and often not fully applicable in practical situations \cite{liu2015low,xu2016robust}. Compared to the original matrix, a low-rank matrix inevitably loses some high-order structural information. In addition,  the consistency between the low-rank matrix and the original high-order matrix is also difficult to maintain because of the structural complexity of the original high-dimensional data space. 
\begin{figure*}[t]
\centering
\includegraphics[width=0.9\linewidth]{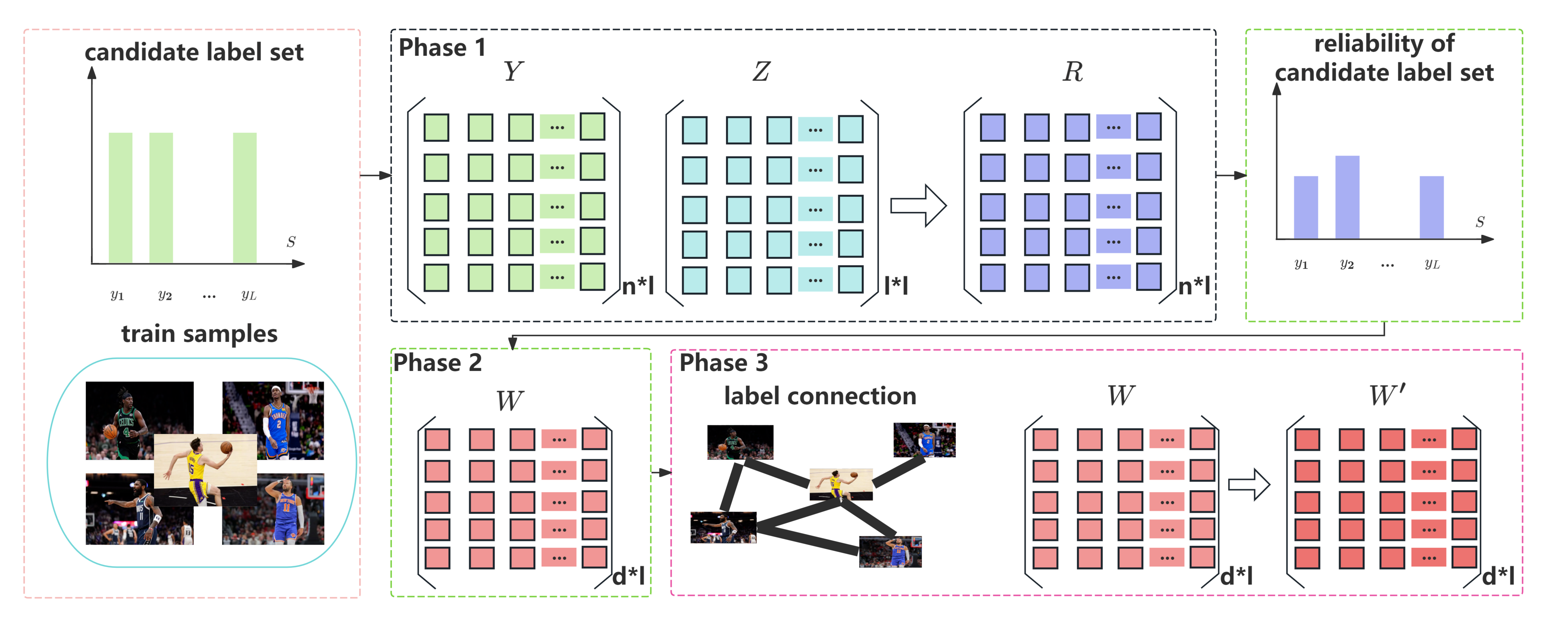}

\caption{Illustration of PML-FSMIR. In the first stage, label matrix is reconstructed with mutual information matrix to get the reliability of the candidate label set.  We reform low-rank assumption to avoid the potential issues in the second stage. Finally in the third stage, weight matrix is reconstructed with the label connection to find the representative labels.  }
\label{Figure 1}
\end{figure*}

Another challenge arises from the inherent sparsity of positive labels within multi-label datasets \cite{liu2021emerging}. This characteristic of datasets poses a significant hurdle, diminishing the efficacy of many multi-label methods in  identifying positive labels due to the lack of samples with positive labels. Consequently, the performance of the methods in recognizing positive labels tends to be suboptimal. However, accurate identification of positive labels is important in prediction tasks in various domains such as disease diagnosis \cite{pham2022graph}, sentiment analysis \cite{yilmaz2021multi}, and gene detection \cite{wang2020deep}. Thus it is necessary for methods capable of  addressing this challenge. The sparsity of positive labels demands that we make the most of the available information. Feature selection can enhance the representativeness of the selected features with respect to positive labels, making it easier for the model to capture the underlying pattern.

To address these two challenges, we propose a novel PML feature selection framework 
based on two crucial characteristics of the datasets. The first one is the noise-resistance of the label relationship: The  relationship of the label space is relatively stable and won't be affected by small amount of noise in it. This is because a group-level perspective can smooth out anomalies or noises in individual data points, thereby providing more stable and trustworthy information. Leveraging the relationship information of label space can effectively reduce noisy labels. Another is the label connectivity of label space: Labels in the dataset are not isolated entities but part of a network of interrelations that can significantly influence their importance in the dataset. In this network, the importance of a label is associated with the number of labels it is connected to and the intensity of their connections. The more numerous and stronger these connections, the higher the perceived importance of the labels. If the model can accurately identify an important label, then the model could also identify the highly connected labels accurately. Thus identifying these important labels can effectively enhance the model's precision.

To make use of these two characteristics, mutual information  is introduced to measure this  structural stability as  it can effectively quantifies both linear and nonlinear relationships between variables \cite{shannon1948mathematical}. We propose a Partial Multi-label Feature Selection model based on Mutual Information Reconstruction (PML-FSMIR). The whole process is illustrated in Figure \ref{Figure 1}. Consequently, the of noise-resistance of the label relationship is utilized to reconstruct the label set, numerical values derived from mutual information matrix  are employed to represent the reliability of candidate labels. Subsequently, the reconstructed label set is adopted as input for model learning, ensuring that the dimensions of the feature and label spaces remain unchanged throughout the learning process at the same time.
Then with the label connectivity, this method reconstructs the learned weight matrix through mutual information matrix. Through this step,  we identify important labels and increase their weights to enhance the performance of the model. Experimental validations on datasets from diverse domains affirm the method's efficacy. Our main contributions are:
\begin{itemize}
    \item We have creatively proposed a method that couples mutual information and sparse learning to make use of collective label characteristics. 
    \item This method breaks free the low-rank assumption commonly used in the past for PML, maintaining the dimensions of the sample space and preserving the high-dimensional information within it.
    \item Extensive experiments have been conducted on datasets in different fields, and the experimental results have demonstrated the superiority of the model.
\end{itemize}
\section{Related Work }
\subsection{Multi-label Feature Selection }
Over the past years, many traditional multi-label feature selection methods have been proposed. They can be broadly categorized into three main types: filters, wrappers and embeddings.
Filter methods select important features by designing a metric, and this process is independent of the subsequent task. Some methods employ mutual information as the metric \cite{gao2018class,zhou2022feature}. MDMR combines mutual information with a max-dependency and min-redundancy method to select superior feature subset for multi-label learning \cite{lin2015multi}.  Others utilize structural similarity as the metric \cite{zhang2021conditional,zhang2021feature,li2024multi}. Others design new evaluation criteria based on existing metrics. Zhang et. al. propose a method that distinguishes three types of label relationships (independence, redundancy and supplementation) and considers changes of label relationships based on different features \cite{zhang2021multi}. GMM proposes  geometric mean to aggregate the mutual information of multiple labels \cite{gonzalez2020distributed}.

Wrapper methods evaluate the performance of feature subsets by training a specific model \cite{chandra2021survey,rigatti2017random}. Due to their time-consuming nature and the need for specific model, they are not used as comparative methods in this paper. 

Embedding methods assess features by designing specific models and integrate
the process of feature selection with learning method.  A typical idea is to design a function to carry out the process of learning and feature selection at the same time and then optimize this function \cite{li2023robust}. MIFS decomposes multi-label information into a low-dimensional space and then employ the reduced space to steer feature selection process \cite{jian2016multi}. Hu et. al. propose a method utilizes Coupled Matrix Factorization (CMF) to extract the shared common mode between feature matrix and label matrix, considering the comprehensive data information in  two matrices \cite{hu2020multi}. 

However, the above-mentioned works are unable to select optimal features in partially multi-label datasets as they are built on the assumption that all observed labels are correct.

\subsection{Partial Multi-label learning }
Partial Multi-label Learning (PML) addresses the issue in which each example is assigned to a candidate label set, and only a part of them is considered correct. To handle PML problem, one approach is to treat the candidate set as ground truth labels directly so that multi-label learning methods can be applied in dealing with PML problem. Nevertheless, these methods are susceptible to being misled by false positive labels concealed within the candidate label set. Hence, there have been proposals for methods specifically tailored to address the challenges of PML. Two effective methods PML-LC and
PML-FP are first proposed by estimating a confidence value for each candidate
label and training a classifier by optimizing the label ranking
confidence matrix \cite{xie2018partial}. These  methods identify noisy labels by the relationship between labels and features \cite{li2021partial}. Yu et. al. propose a method utilizes a low-rank matrix approximation and latent dependencies between labels and features to identify noisy labels and train a multi-label classifier \cite{yu2018feature}. PARTICAL-MAP and
PARTICAL-VLS elicit credible labels from the candidate label set for model induction \cite{zhang2020partial}.  From  the other view, Xu et. al.  propose a method leverages the topological information of the feature space and the correlations among the labels to recover label distributions \cite{xu2020partial}. Another view is how to exploit the negative information like the non-candidate set. Li and Wang propose a method that recovers the ground-truth labels by estimating the ground-truth confidences from the label enrichment, which is composed of the relevance degrees of candidate labels and irrelevance degrees of non-candidate labels \cite{li2020recovering}. PML-LFC estimates the confidence values of relevant labels for each instance using the similarity from both the label and feature spaces, and trains the desired predictor with the estimated confidence values \cite{yu2020partial}. 
 
 However, the above methods haven't considered the problem of sparse positive labels in the dataset. In order to strengthen the identification ability of positive labels, we propose PML-FSMIR.

\section{The Proposed Method }
The proposed method consists of three stages: (1) label matrix reconstruction stage aims to reduce noises in labels; (2) reformed low-rank assumption stage aims to exploit cleaned information for feature selection with reformed low-rank assumption; (3) weight matrix reconstruction stage aims to enhance sensitivity to key labels for final feature selection.
\subsection{Label Matrix Reconstruction }
In the first stage, to reduce the influence of noises in the label set, the label mutual information matrix is introduced to reconstruct the label matrix. Label relationship is more stable than a single label, which means it is less likely to be affected by the noises in the label set. Thus  label mutual information matrix which reflects the degree of the label relationship is adopted to 
reduce the noises.

The input consists of two parts: the feature matrix \( X=\left[x_1, x_2 , \ldots x_n\right] \in R^{n \times d}\) where \( d \) is the dimension of feature vector and \( n \) is the number of training instances and label matrix 
\( Y=\left[y_1, y_2 , \ldots y_n\right] \in [0,1]^{n \times q}\), where \( q \) is the dimension of label vector. In label matrix, \( y_{i j}=1 \) means that the \( j \)-th label is a candidate label of the \( i \)-th instance.

Initially,  we compute the mutual information between labels and form them into a \( q \)-dimensional square matrix \( Z \):
\begin{equation}
    Z_{i j}=I\left(y_{:i}, y_{:j}\right).
\end{equation}%
\( Z_{i j} \) denotes the mutual information between \( i \)-th and \( j \)-th label. As mutual information 
can quantify  relationship between two variables and structure of the label is more stable and less likely to be influenced by noises in single sample, the mutual information between labels 
can be employed with candidate labels to determine the reliability of candidate labels. 
Then we can introduce the label reconstruction matrix \( T \):

\begin{equation}
\begin{aligned}
& T=(Y Z) \circ \operatorname{sign}(Y), \\
& \operatorname{sign}\left(y_{i j}\right) = 
\begin{cases} 
0, & y_{i j}=0 \\ 
1, & y_{i j}=1 
\end{cases}
\label{f2}
\end{aligned}
\end{equation}
Simply using the dot product of \( Y \) and \( Z \) would causes  non-candidate labels to be assigned with non-zero values,  the  \( sign(Y) \) is further employed so that non-candidate labels will remain as zero. In this matrix, \( T_{i j} \) can be expressed as:
\begin{equation}
 T_{i j}=y_{ij}\sum_{k=1}^q y_{i k} I\left(y_{i j}, y_{i k}\right)
\end{equation}

Compared to \( y_{ij} \), \( T_{ij} \) comprehensively considers the relationship between labels and the candidate label set in the sample: the more labels in the candidate label set that are highly associated with it, the greater the value. For  candidate label \( y_{ij} \), if the labels that are structurally similar to it in the label space are included in the candidate set, then \( T_{ij} \)  will be large, indicating that the possibility of it being noise is low. Conversely, if the labels similar to it are not in the candidate set, then \( T_{ij} \)  will be small, suggesting that it is very likely to be noise. 
After normalization, this matrix \( T \) can replace the original matrix \( Y \) to the subsequent stages. The effectiveness of this step would be 
demonstrated in ablation experiments.

\subsection{Reformed Low-rank Assumption }
In the second stage, we hope to construct the objective function while avoiding the possible problems of the traditional low-rank assumption. For the label matrix \( Y \) , we have reconstructed the label matrix \( T \)  in the first stage for disambiguation, thus no operations is needed in this stage. 
However,  the noises and redundancy issues that exist in \( X \) still exist. So we adpot the reformed low-rank assumption to simultaneously remove noises and avoid redundancy and potential issues in low-rank assumption:
\begin{equation}
    \min _{U, V, W}\|UVW-T\|_F^2+\alpha\|X-U V\|_F^2.
\end{equation}
Where \( U \in R^{n \times k}\), \( V \in R^{k \times d}\) and \( W \in R^{d \times q}\) represent cluster matrix, cluster weight matrix, and feature weight matrix respectively. The traditional low-rank assumption uses matrix decomposition to remove redundancy and noises in the original matrix. We retain this matrix decomposition term, but adopt \( UV \) instead of the decomposed \( U \) to replace the original \( X \). The dimension of \( X \) is preserved, and  the goal of removing noises is achieved without losing the high-dimensional structural information of \( X \). In this way, the low-rank assumption is reformed.

To ensure the weight matrix \(W\) have the same structure as the original data points. We apply a manifold regularization term to \(W\). Specifically, the  mutual information matrix \(Z^{\prime} \) of \( T \) is employed
to replace the affinity matrix as this matrix is less sensitive to noises. It can be expressed as:
\begin{equation}
    Z^{\prime}_{i j}=I\left(T_{:i}, T_{:j}\right).
\end{equation}%
So the formula is changed as follow:

\begin{equation}
    \min _{U, V, W}\|UVW-T\|_F^2+\alpha\|X-U V\|_F^2+\beta \operatorname{Tr}(W) L_T(W)^T.
    \label{f5}
    \end{equation}
Where \(L_T = A- Z^{\prime}\) is graph laplacian matrix of \(T\) and \(A\) is a diagonal matrix. Finally, to achieve feature selection, we further add a \( l_{2,1} \)-norm of \(W\) to Formula \ref{f5}  \cite{nie2010efficient}:
\begin{equation}
\begin{aligned}
    &\min _{U, V, W}\|UVW-T\|_F^2+\alpha\|X-U V\|_F^2+\beta \operatorname{Tr}(W) L_T(W)^T \\&+\gamma\|W\|_{2,1}.
     \end{aligned}
\end{equation}

For optimization, this formula involves three variables \(W\), \(U\) and \(V\). After relax the \( W_{2,1}\) into  \(\operatorname{Tr}(W)^T Q(W)\) where \(Q\) is a diagonal matrix: \(Q_{ii}=\frac{1}{2 \sqrt{W_i^T W_i+\epsilon}}, (\epsilon \rightarrow 0).\) The objective function can be rewritten as:
\begin{equation}
    \begin{aligned}
    &\Theta(U,V,W)=\operatorname{Tr}\left((UVW-T)^T(UVW-T)\right)+
    \\&\alpha \operatorname{Tr}\left((X-UV)^T(X-UV)\right) +\beta\operatorname{Tr}W L_T W^T+\\&\gamma \operatorname{Tr}\left(W^T Q W\right). \label{f7}
    \end{aligned}
\end{equation}
Multiplicative gradient descent strategy is adopted to solve Formula \ref{f7}  \cite{beck2009fast}, in each iteration,  each variable is updated while fixing other variables. By taking derivative of Formula \ref{f7} based on KKT conditions, we have:
\begin{equation}
    U_{i j}^{t+1} \leftarrow U_{i j}^t \frac{\left( T W^T V^T+\alpha XV^T\right)_{i j}}{\left(UVWW^T V^T+\alpha  UVV^T\right)_{i j}}.\label{f8}
    \end{equation} 
\begin{equation}
    V_{i j}^{t+1} \leftarrow V_{i j}^t \frac{\left( U^TT W^T +\alpha U^TX\right)_{i j}}{\left(U^TUVWW^T+\alpha  U^TUV\right)_{i j}}. \label{f9}
\end{equation}
\begin{equation}
    W_{i j}^{t+1} \leftarrow W_{i j}^t \frac{\left(V^TU^TT\right)_{i j}}{\left(V^TU^TUV W+\beta W+\delta Q W\right)_{i j}}.\label{f10}
\end{equation}
In real implementation, a small positive value is added to the denominator to avoid zero values.
\begin{algorithm}[h]
    \caption{Pseudo code of PML-FSMIR}
    \label{alg:method}
    \textbf{Input}: Feature matrix \(X\) and label matrix \(Y\), regularization parameters \(\alpha\), \(\beta\), and \(\gamma\).\\
    \textbf{Output}: Return the ranked features.
    
    \begin{algorithmic}[1] 
        \STATE Construct mutual information matrix \(Z\) of \(Y\);
        \STATE Calculate \(T\)  by Formula \ref{f2};
        \STATE Construct mutual information matrix \(Z^{\prime}\) and graph laplacian matrix of \(T\); 
        \WHILE{not coverage}
        \STATE Calculate \(Q\);
        \STATE Update \(U\) by Formula \ref{f8} with other variables fixed;
        \STATE Update \(V\)  by Formula \ref{f9} with other variables fixed;
        \STATE Update \(W\)  by Formula \ref{f10} with other variables fixed;
        \ENDWHILE
        \STATE Reconstruct \(W\) by Formula \ref{f11};
        \STATE \textbf{return} Return features according to \(\left\|W_i.\right\|_2\).
    \end{algorithmic}
\end{algorithm}
\begin{figure}[t]
\centering
\includegraphics[width=0.9\columnwidth]{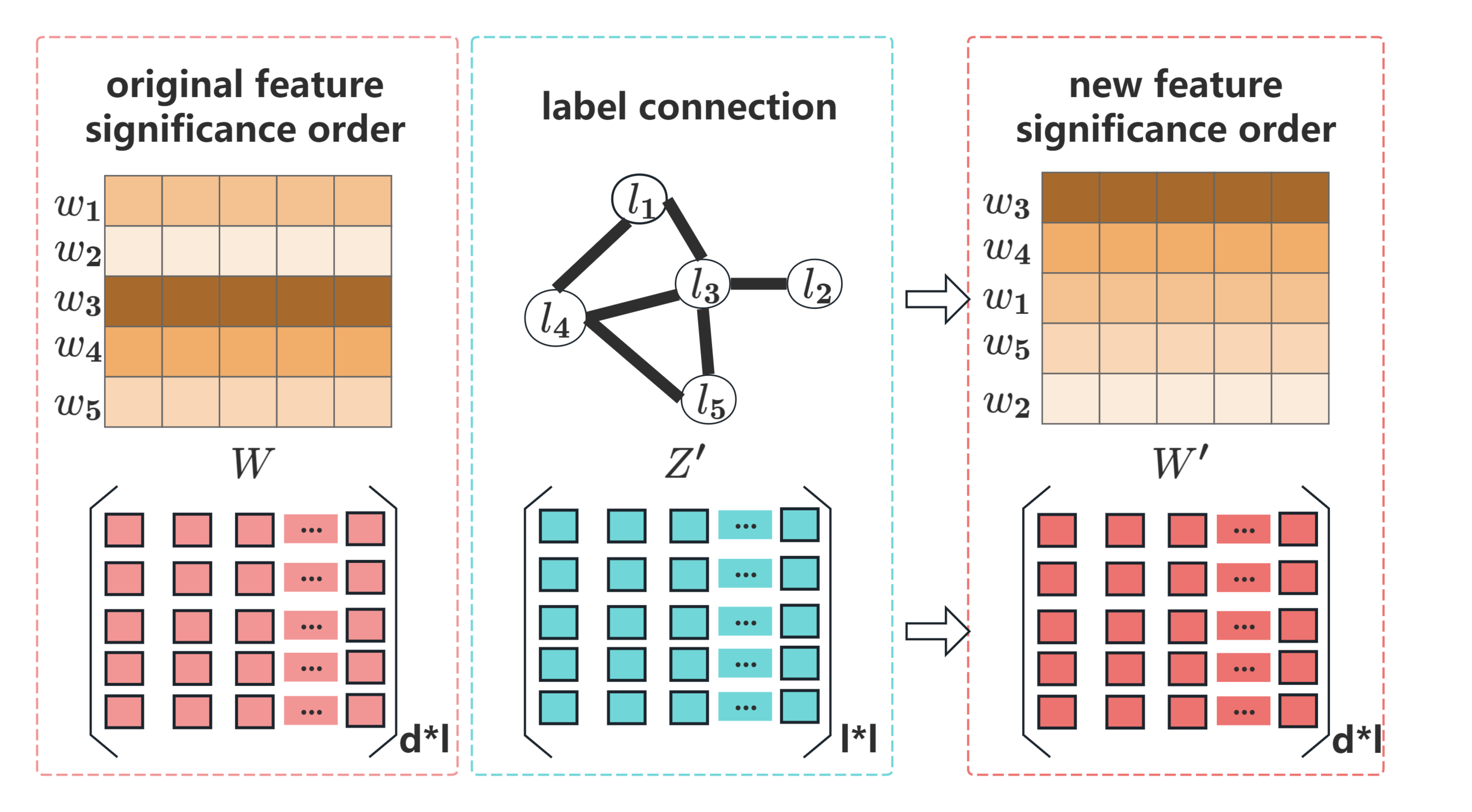} 
\caption{The weight matrix \(W\) is reconstructed through the mutual information matrix \(Z'\) representing the label connection, and the correct feature weight is obtained.}
\label{Figure 2}
\end{figure}
\subsection{Weight Matrix Reconstruction}
In partial multi-label datasets, positive labels are often more sparse but more important than negative labels \cite{liu2006semi}. To further enhance the predicting ability of selected features of positive labels, we design this stage using  mutual information matrix \(Z^{\prime} \) to reconstruct the weight matrix.

To this end, the keypoint is to identify some representative labels which are related to more labels compared to other labels. If we can find those labels that are highly connected with a large number of labels and select these features that can identify these labels, the experimental results will significantly improve. As shown in Figure \ref{Figure 2}, since features is considered as equal after the end of  the second stage, the feature weights are not correctly ranked as expected. Therefore,  the weight matrix needs to be reconstructed using label connection, so that the reconstructed weight matrix can better identify the features that are more important to the representative labels. In weight matrix \(W\), \( W_{i j} \) is the weight of the \(i\)-th feature to the \(j\)-th label, so the sum of the row \( W_{i} \) can be regarded as the importance of the \(i\)-th feature to the label set.
To enhance the identification ability of the selected feature to the representative labels, we utilize the mutual information between labels to reconstruct \(W\):
\begin{equation}
    W_{i j}=\sum_{k=1}^q W_{i k} Z^{\prime}_{k j}.\label{f11}
\end{equation}
Through this equation, \( W_{i j} \), the weight of \(i\)-th feature to \(j\)-th label is updated according to the relevance of this labels to other labels. After this 
operation, those features that have higher weight to representative labels would be selected. According to final
value of \(\left\|W_{i \cdot}\right\|_2(i=1, \ldots, d)\) in a descending order,  the top ranked features are obtained. The pseudo code is presented in Algorithm \ref{alg:method}.
Subsequent ablation experiments have demonstrated the necessity of this stage.

\section{Experiments}

\subsection{Datasets}
We perform experiments on eight datasets from a broad range of applications: \textbf{\textit{Birds}} for audio, \textbf{\textit{CAL500}} for music classification, \textbf{\textit{Corel5K}}  for image
annotation, \textbf{\textit{LLOG\_F}} and \textbf{\textit{Slashdot}} for  text categorization, \textbf{\textit{Water}} for chemistry, \textbf{\textit{Yeast}} for gene function prediction, and \textbf{\textit{CHD49}}
for medicine. We keep the noisy level of every dataset at 20\%. On each dataset, ten-fold cross-validation is performed where the mean metric values as well as standard deviations are recorded for each compared method. Detailed information is shown in Table \ref{t1}.
\begin{table}[h]
    \resizebox{\linewidth}{!}{%
    \begin{tabular}{@{}llllllllll@{}}
    \toprule
    Name     & Domain    & \#Instances & \#Features & \#Labels \\ \midrule
    Birds \cite{briggs20139th}    & audio     & 645        & 260        & 19       \\
    CAL \cite{turnbull2008semantic}      & music     & 555        & 49         & 6        \\
    CHD\_49 \cite{shao2013symptom}  & medicine  & 555        & 49         & 6        \\
    Corel5K \cite{duygulu2002object} & image     & 5000       & 499        & 374      \\
    LLOG\_F \cite{read2010scalable} & text      & 1460       & 1004       & 75       \\
    Slashdot \cite{read2010scalable} & text      & 3782       & 1079       & 22       \\ 
    Water \cite{blockeel1999simultaneous}     & chemistry & 1060       & 16         & 14       \\
    Yeast  \cite{elisseeff2001kernel}  & biology   & 2417       & 103        & 14       \\\bottomrule
    \end{tabular}
    }
    
    \caption{ Characteristics of experimental datasets.}
    \label{t1}
\end{table}
\begin{table*}[htbp]
\centering
    \resizebox{0.95\linewidth}{!}{%
    \begin{tabular}{@{}llllllllll@{}}
    \toprule
    Datasets & PML-FSMIR               & PML-LC     & PML-FP     & PAR-VLS    & PAR-MAP    & FPML        & PML-FSSO       & MIFS        & DRMFS       \\\midrule
    Birds    & \textbf{0.32±0.09} & 0.47±0.15 & 0.38±0.15 & 0.59±0.12 & 0.69±0.05  & 1.00±0.00         & 0.33±0.08 & 0.73±0.19 & 0.48±0.23 \\
    CAL      & \textbf{0.31±0.07} & 0.63±0.12  & 0.61±0.15 & 0.69±0.13 & 0.72±0.07 & 0.68±0.09 & 0.39±0.10 & 0.58±0.26 & 0.57±0.17 \\
    CHD\_49  & \textbf{0.29±0.11} & 0.46±0.16 & 0.47±0.13  & 0.34±0.14 & 0.61±0.06 & 0.72±0.08  & 0.31±0.19 & 0.35±0.12 & 0.33±0.17 \\
    Corel5K  & \textbf{0.49±0.19} & 0.65±0.12 & 0.64±0.12 & 0.84±0.08 & 0.90±0.06 & 0.92±0.02 & 0.63±0.019 & 0.76±0.15 & 0.71±0.16  \\
    LLOG\_F   & \textbf{0.31±0.10} & 0.76±0.02          & 0.69±0.02       & 0.81±0.05 & 0.67±0.11 & 0.53±0.18 & 0.45±0.11  & 0.50±0.21 & 0.63±0.20 \\
    Slashdot & \textbf{0.05±0.04} & 0.49±0.22  & 0.47±0.21 & 0.44±0.27 & 0.43±0.27 & 1.00±0.00         & 0.05±0.07  & 0.34±0.25 & 0.60±0.23 \\
    Water    & \textbf{0.36±0.01} & 0.46±0.06 & 0.44±0.06 & 0.48±0.03 & 0.44±0.03  & 0.44±0.03 & 0.39±0.06 & 0.44±0.03 & 0.42±0.07 \\
    Yeast    & \textbf{0.24±0.09}  & 0.43±0.01          & 0.45±0.02       & 0.34±0.14 & 0.62±0.01 & 0.34±0.12 & 0.35±0.14 & 0.24±0.10 & 0.37±0.16\\ \bottomrule
    \end{tabular}
    }
    \caption{ Experimental results (mean ± std) in terms of Ranking Loss where the best performance is shown in boldface.}
    \label{t2}
    \end{table*}

\begin{table*}[htbp]
\centering
        \resizebox{0.95\linewidth}{!}{%
        \begin{tabular}{@{}llllllllll@{}}
        \toprule
        Datasets  & PML-FSMIR               & PML-LC     & PML-FP     & PAR-VLS    & PAR-MAP     & FPML        & PML-FSSO       & MIFS        & DRMFS                \\ \midrule
        Birds     & \textbf{0.47±0.05} & 0.61±0.07 & 0.55±0.08  & 0.57±0.09 & 0.62±0.04 & 0.87±0.00     & 0.48±0.06 & 0.74±0.10 & 0.56±0.14          \\
        CAL       & \textbf{0.63±0.02}  & 0.81±0.01 & 0.79±0.01 & 0.74±0.03 & 0.71±0.02 & 0.71±0.03 & 0.66±0.04 & 0.74±0.06 & 0.72±0.04          \\
        CHD\_49   & 0.50±0.04          & 0.62±0.04  & 0.62±0.03 & 0.66±0.02 & 0.66±0.02 & 0.67±0.01 & 0.49±0.07 & 0.53±0.04 & \textbf{0.49±0.07} \\
        Corel5K   & \textbf{0.37±0.08} & 0.51±0.03  & 0.48±0.03 & 0.55±0.03 & 0.59±0.01 & 0.57±0.01 & 0.43±0.08 & 0.51±0.06   & 0.47±0.07          \\
        LLOG\_F    & \textbf{0.53±0.01}  & 0.51±0.03    & 0.48±0.03       & 0.59±0.00 & 0.56±0.01 & 0.55±0.02 & 0.59±0.02 & 0.57±0.02 & 0.59±0.00          \\
        Slashdot  & \textbf{0.06±0.01} & 0.23±0.06  & 0.23±0.06 & 0.19±0.09 & 0.19±0.08  & 0.37±0.00     & 0.07±0.02 & 0.16±0.08 & 0.24±0.08          \\
        Water     & \textbf{0.72±0.01} & 0.76±0.01  & 0.75±0.01 & 0.73±0.00 & 0.73±0.01 & 0.74±0.00 & 0.73±0.04 & 0.74±0.01 & 0.75±0.03          \\
        Yeast     & \textbf{0.50±0.03} & 0.56±0.08    & 0.58±0.02        & 0.56±0.05 & 0.78±0.00     & 0.53±0.04 & 0.58±0.05 & 0.52±0.07 & 0.58±0.06          \\ \bottomrule
        \end{tabular}
        }
        \caption{ Experimental results (mean ± std) in terms of Coverage where the best performance is shown in boldface.}
        \label{t3}
\end{table*}

\begin{table*}[htbp]
\centering
    \resizebox{0.95\linewidth}{!}{%
    \begin{tabular}{@{}llllllllll@{}}
    \toprule
    Datasets & PML-FSMIR               & PML-LC     & PML-FP     & PAR-VLS    & PAR-MAP     & FPML        & PML-FSSO       & MIFS        & DRMFS       \\\midrule
    Birds    & \textbf{0.59±0.03}  & 0.36±0.06 & 0.49±0.08  & 0.49±0.10 & 0.46±0.05 & 0.16±0.00     & 0.54±0.05  & 0.31±0.10 & 0.43±0.04 \\
    CAL      & \textbf{0.59±0.03}  & 0.39±0.01 & 0.41±0.01 & 0.48±0.05 & 0.55±0.04  & 0.55±0.05 & 0.51±0.04 & 0.47±0.08 & 0.43±0.04 \\
    CHD\_49  & \textbf{0.77±0.02} & 0.68±0.01 & 0.66±0.01 & 0.76±0.02 & 0.71±0.01 & 0.65±0.02  & 0.77±0.04 & 0.75±0.02 & 0.77±0.03 \\
    Corel5K  & \textbf{0.42±0.08} & 0.30±0.03 & 0.32±0.04 & 0.29±0.04  & 0.24±0.02  & 0.27±0.02 & 0.37±0.08 & 0.28±0.05 & 0.33±0.07 \\
    LLOG\_F   & 0.59±0.02 & 0.47±0.01   & 0.48±0.01       & 0.48±0.01 & 0.57±0.03 & \textbf{0.60±0.04}   & 0.46±0.01 & 0.54±0.03 & 0.47±0.00 \\
    Slashdot & \textbf{0.93±0.03} & 0.42±0.11 & 0.43±0.11 & 0.64±0.23  & 0.65±0.22 & 0.17±0.00     & 0.93±0.05 & 0.70±0.19 & 0.49±0.18 \\
    Water    & 0.60±0.01 & 0.52±0.01 & 0.53±0.01  & 0.60±0.01 & 0.59±0.02 & \textbf{0.61±0.02} & 0.58±0.04 & 0.53±0.03 & 0.55±0.05 \\
    Yeast    & \textbf{0.70±0.04} & 0.51±0.03   & 0.52±0.03       & 0.61±0.04 & 0.57±0.01  & 0.63±0.04 & 0.57±0.04  & 0.68±0.08 & 0.58±0.05 \\ \bottomrule
    \end{tabular}
    }

    \caption{ Experimental results (mean ± std) in terms of Average Precision where the best performance is shown in boldface.}
    \label{t4}
\end{table*}

\begin{table*}[htbp]
\centering
    \resizebox{0.95\linewidth}{!}{%
    \begin{tabular}{@{}llllllllll@{}}
    \toprule
    Datasets   & PML-FSMIR               & PML-LC     & PML-FP     & PAR-VLS    & PAR-MAP     & FPML        & PML-FSSO     & MIFS        & DRMFS       \\\midrule
    Birds      & \textbf{0.38±0.04} & 0.11±0.04 & 0.26±0.08 & 0.19±0.01 & 0.28±0.08 & 0.00±0.00         & 0.24±0.03 & 0.01±0.07 & 0.11±0.05 \\
    CAL        & \textbf{0.49±0.08} & 0.02±0.03 & 0.07±0.06 & 0.02±0.00     & 0.08±0.08 & 0.10±0.06 & 0.00±0.00 & 0.00±0.00          & 0.00±0.00         \\
    CHD\_49    & \textbf{0.56±0.19} & 0.27±0.13 & 0.23±0.11 & 0.16±0.12 & 0.24±0.12 & 0.01±0.01 & 0.19±0.13 & 0.13±0.10 & 0.08±0.11 \\
    Corel5K    & \textbf{0.13±0.08} & 0.06±0.02 & 0.05±0.03 & 0.04±0.04 & 0.03±0.01  & 0.00±0.00         & 0.00±0.00         & 0.00±0.00          & 0.02±0.02 \\
    LLOG\_F     & \textbf{0.32±0.03} & 0.00±0.00        & 0.00±0.00         & 0.03±0.00     & 0.23±0.11 & 0.22±0.06 & 0.10±0.04   & 0.15±0.06  & 0.02±0.01 \\
    Slashdot   & \textbf{0.76±0.10} & 0.21±0.10 & 0.14±0.10 & 0.34±0.21 & 0.37±0.20 & 0.00±0.00          & 0.00±0.00 & 0.13±0.09 & 0.05±0.05 \\
    Water      & \textbf{0.48±0.11}  & 0.32±0.15 & 0.34±0.17 & 0.10±0.06 & 0.17±0.11 & 0.11±0.07 & 0.13±0.09 & 0.44±0.02 & 0.07±0.09 \\
    Yeast      & \textbf{0.46±0.16}   &0.00±0.00        & 0.00±0.00      & 0.00±0.00          & 0.00±0.00          & 0.03±0.03          & 0.02±0.03 & 0.20±0.15  & 0.03±0.01\\ \bottomrule
    \end{tabular}
    }
    
    \caption{ Experimental results (mean ± std) in terms of Marco-F1 where the best performance is shown in boldface.}
    \label{t5}
\end{table*}

\begin{table*}[htbp]
\centering
\resizebox{0.95\linewidth}{!}{%
\begin{tabular}{@{}llllllllll@{}}
    \toprule
    Datasets   & PML-FSMIR               & PML-LC     & PML-FP     & PAR-VLS    & PAR-MAP     & FPML        & PML-FSSO     & MIFS        & DRMFS       \\\midrule
Birds      & \textbf{0.38±0.03} & 0.12±0.04 & 0.27±0.08 & 0.20±0.10 & 0.29±0.09 & 0.00±0.00         & 0.33±0.09 & 0.11±0.07 & 0.11±0.05 \\
CAL        & \textbf{0.51±0.06} & 0.02±0.03 & 0.07±0.05 & 0.02±0.00  & 0.08±0.08 & 0.12±0.07  & 0.00±0.00   & 0.00±0.00          & 0.00±0.00         \\
CHD\_49    & \textbf{0.60±0.19} & 0.31±0.15  & 0.24±0.12 & 0.20±0.14 & 0.26±0.12  & 0.01±0.01 & 0.18±0.14 & 0.16±0.12 & 0.09±0.13  \\
Corel5K    & \textbf{0.13±0.08} & 0.08±0.02  & 0.07±0.05 & 0.05±0.06 & 0.04±0.02 & 0.00±0.00         & 0.00±0.00         & 0.00±0.00         & 0.04±0.04 \\
LLOG\_F     & \textbf{0.34±0.02} & 0.00±0.00         & 0.00±0.00        & 0.03±0.00     & 0.23±0.12  & 0.32±0.09 & 0.08±0.04 & 0.19±0.07 & 0.02±0.07 \\
Slashdot   & \textbf{0.81±0.09} & 0.26±0.12 & 0.21±0.14 & 0.40±0.23 & 0.44±0.20 & 0.00±0.00          & 0.00±0.00 & 0.18±0.12 & 0.06±0.07 \\
Water      & \textbf{0.49±0.11} & 0.33±0.15 & 0.35±0.18 & 0.09±0.05 & 0.15±0.11 & 0.11±0.07 & 0.19±0.14 & 0.46±0.02 & 0.06±0.09 \\
Yeast      & \textbf{0.48±0.16}  & 0.00±0.00         & 0.00±0.00        & 0.00±0.00          & 0.00±0.00         & 0.00±0.00         & 0.03±0.03  & 0.23±0.16 & 0.02±0.01 \\\bottomrule
    \end{tabular}
    }
    \label{t6}
    \caption{ Experimental results (mean ± std) in terms of Mirco-F1 where the best performance is shown in boldface.}
\end{table*}

\subsection{Experimental Setup}

Evaluation Metrics: We adopt five widely used multi-label metrics including Ranking Loss, Coverage, Average Precision, Marco-F1, and Micro-F1.
For Ranking Loss and Coverage, the smaller value, the better the performance. For Average Precision, Marco-F1, and Micro-F1, the larger
the value, the better the performance.

Baselines: We implement eight state-of-the-art methods of multi-label learning and PML for comparison and record the feature selection result with twenty points of different percentages. They are one Partial Multi-label Feature Selection methods (PML-FSSO \cite{hao2023partial}), five Partial Multi-label Learning methods (PML-LC \cite{xie2018partial}, PML-FP \cite{xie2018partial}, PAR-VLS \cite{zhang2020partial}, PAR-MAP \cite{zhang2020partial} and FPML \cite{yu2018feature}) and two Multi-label Feature Selection methods (MIFS \cite{2016Multi} and DRMFS\cite{2020Robust}). Due to the lack of feature selection method in partial multi-label learning, the weight matrix is extracted from
model to reflect the importance of features. We adopt ten-fold cross-validation to train these models and the selected
features are compared on SVM classifier\footnote{The code is available at https://github.com/typsdfgh/PML-FSMIR}.

\begin{figure}[htbp]
	\centering
    \begin{subfigure}{\linewidth}
		\centering
		\includegraphics[width=\linewidth]{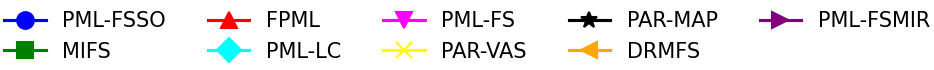}

		\label{ch}
	\end{subfigure}
	\begin{subfigure}{0.49\linewidth}
		\centering
		\includegraphics[width=\linewidth]{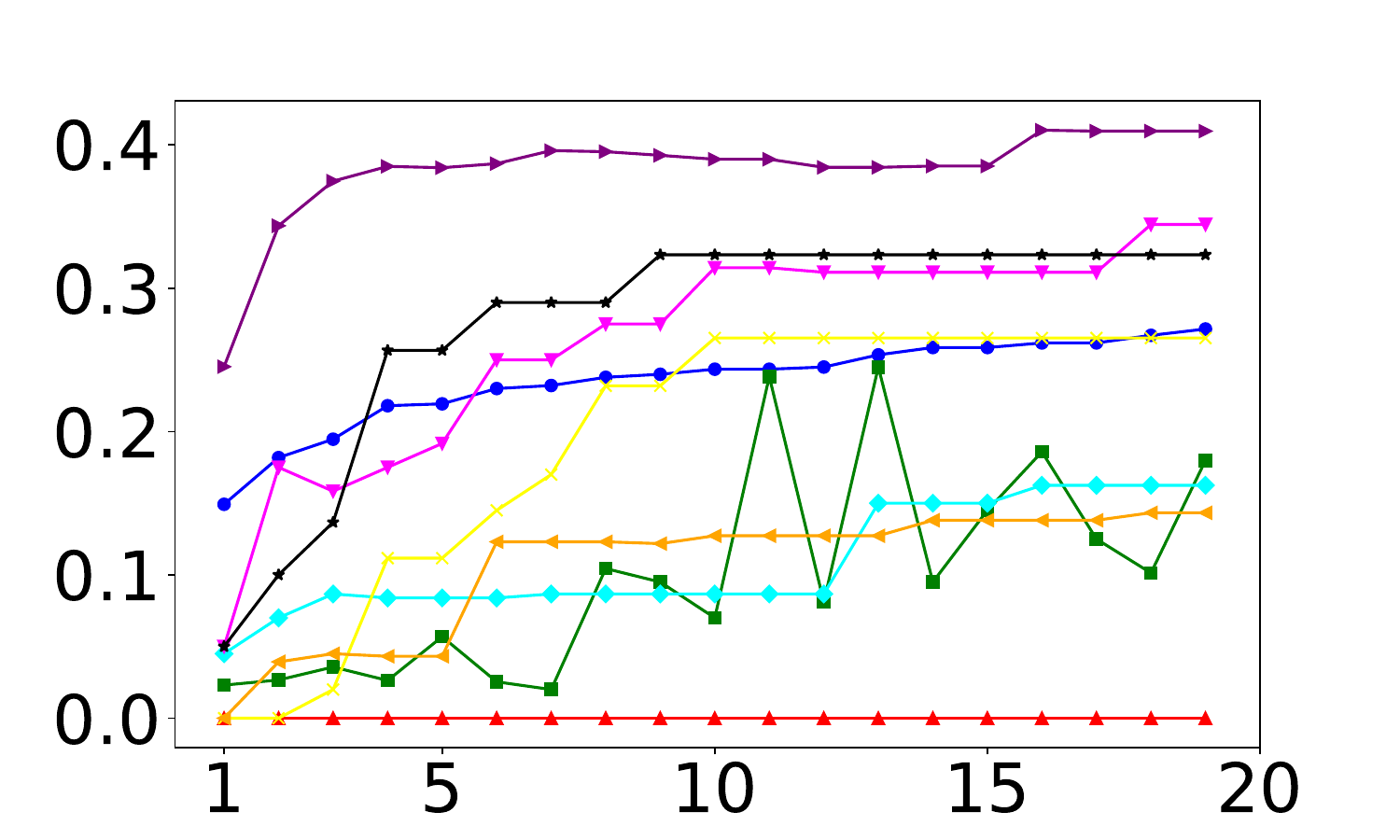}
		\caption{Marco-F1}
		\label{chutian1}
	\end{subfigure}
	\begin{subfigure}{0.49\linewidth}
		\centering
		\includegraphics[width=\linewidth]{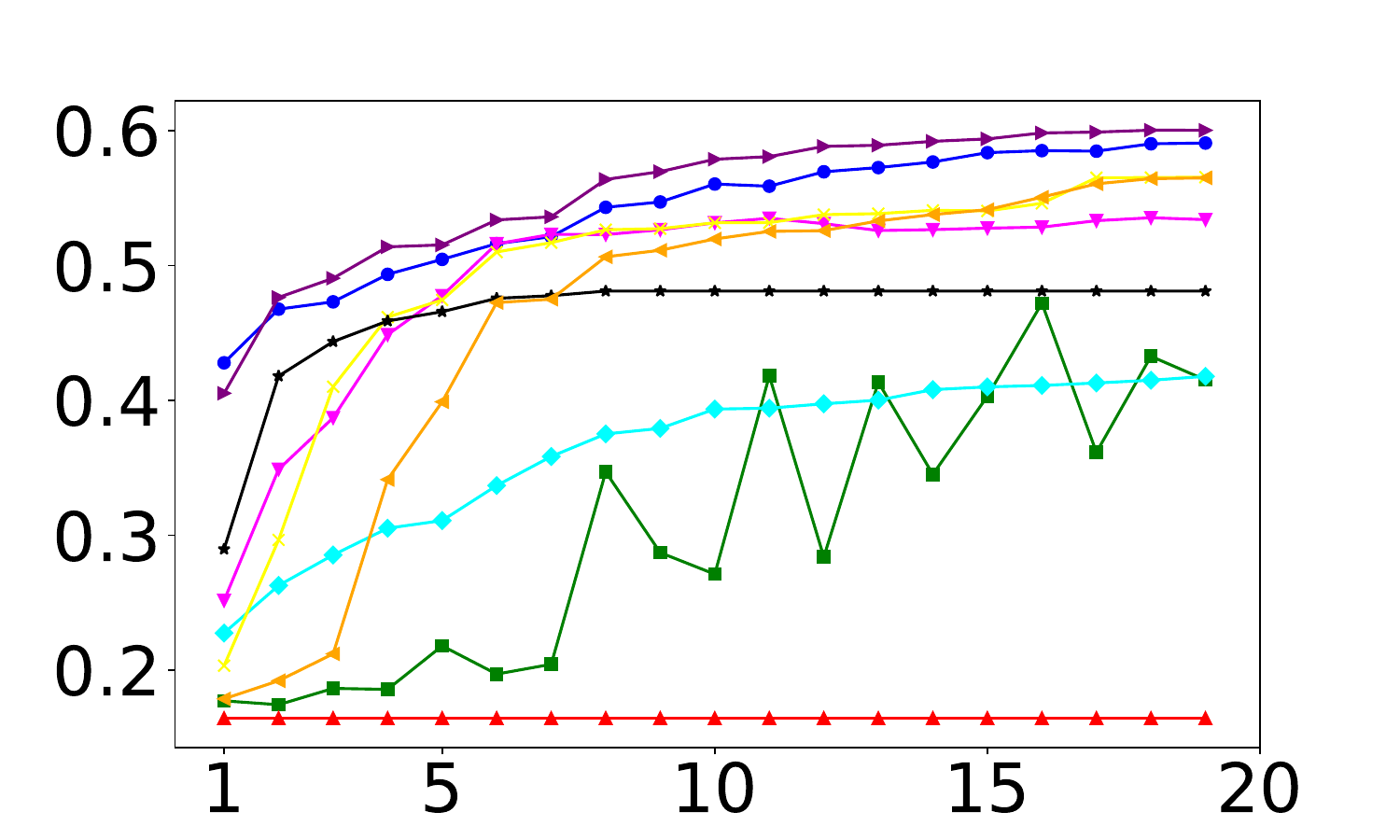}
		\caption{Average
Precision}
		\label{chutian2}
	\end{subfigure}
	
	\begin{subfigure}{0.49\linewidth}
		\centering
		\includegraphics[width=\linewidth]{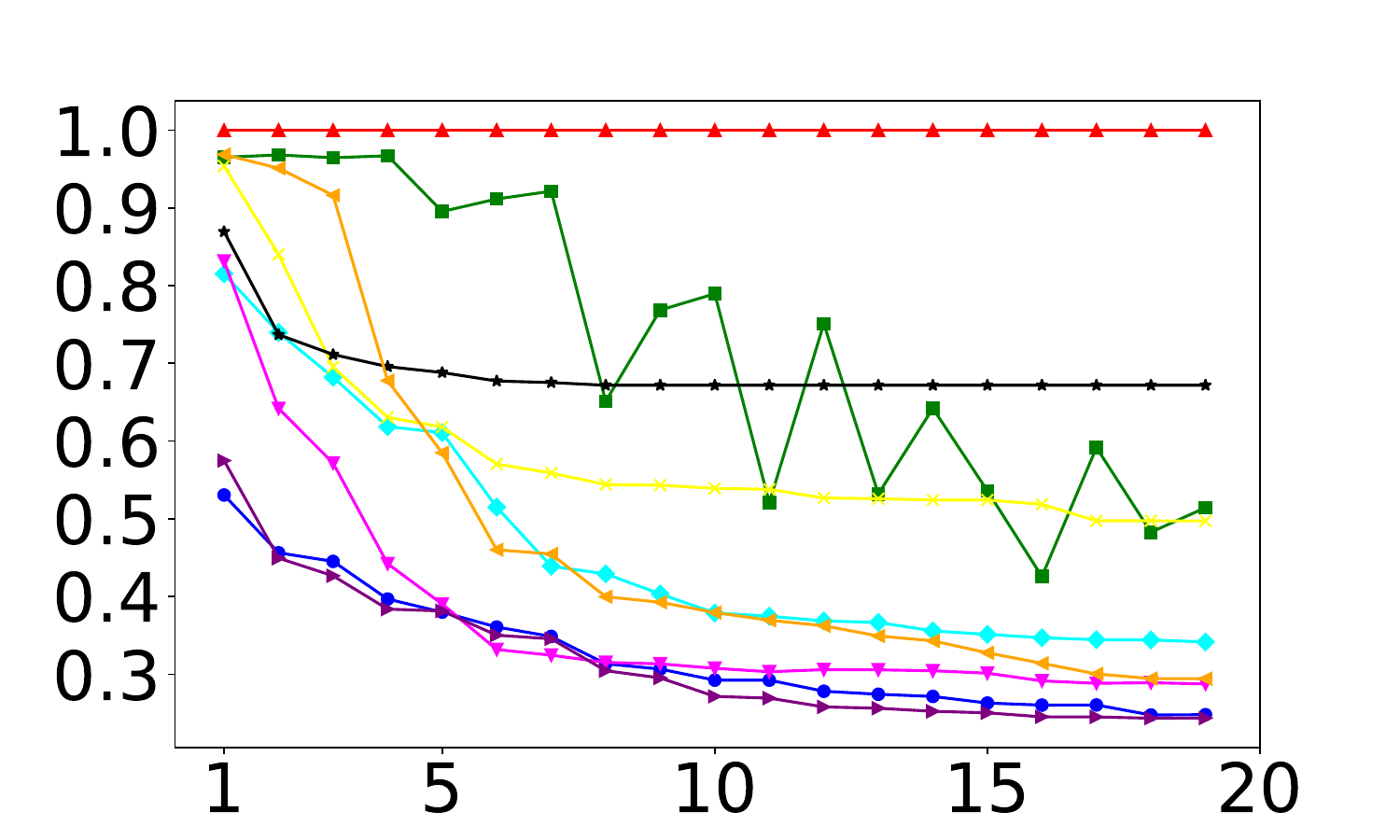}
		\caption{Ranking Loss}
		\label{chutian3}
	\end{subfigure}
	\begin{subfigure}{0.49\linewidth}
		\centering
		\includegraphics[width=\linewidth]{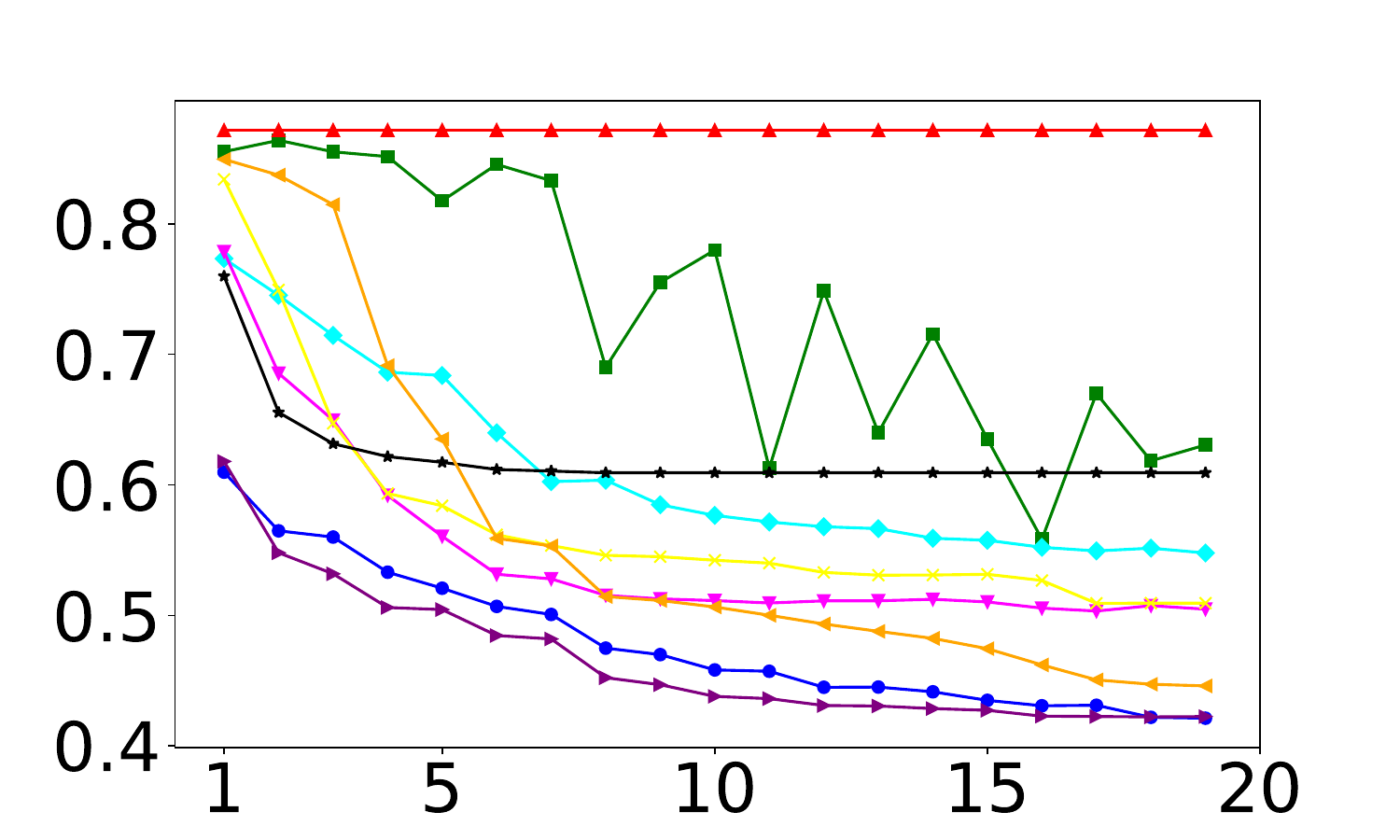}
		\caption{Coverage Error}
		\label{chutian4}
	\end{subfigure}
 \caption{Nine methods on \textbf{\textit{Birds}} in terms of Marco-F1, Average
Precision, Ranking Loss and Coverage..}
	\label{f3}
\end{figure}
\subsection{Results}
The tables below show the detail of the experiment result, for all datasets except \textbf{\textit{Water}},  we select specified number of features according to the importance
as descending order from one to twenty percent for each percentage of the features, the five used metrics for each dataset are recorded in form of mean and standard deviation among different percentages. We also illustrate
one dataset for the detail of all metrics in Figure \ref{curve_bird} to show our performance clearly. As for \textbf{\textit{Water}} only has 16 features, we select 1 to 16 features. From  overall results, we make following observations:
\begin{itemize}
    \item On eight datasets across all evaluation metrics, PML-FSMIR ranks first in all cases except Coverage on \textbf{\textit{CHD\_49}} and  Average Precision on \textbf{\textit{LLOG\_F}}, while in these two cases PML-FSMIR all ranks second. These results fully demonstrate the superiority of PML-FSMIR.
    \item For the superiority in Ranking Loss, Coverage, and Average Precision. We attribute it to the reconstruction of the label matrix through mutual information matrix in the first stage. This step effectively reduces the noises in the labels, making the selected features more helpful for subsequent work. The subsequent ablation experiments further confirmed its function.
    \item For the superiority in Mirco-F1 and Marco-F1. We attribute it to the reconstruction of the weight matrix through mutual information matrix in the third stage. This step improves the weight of features with strong ability to determine key labels by reconstructing the weight matrix through mutual information matrix. The results in Table \ref{t4} and \ref{t5} show that this step significantly improves the F1-score, which improves the best results of the baselines at least by 15.2\%. The subsequent ablation experiments further confirmed this opinion.
\end{itemize}
\begin{figure}[!htbp]
    \begin{subfigure}{0.32\linewidth}
        \centering
        \scalebox{1.3}{\includegraphics[width=\linewidth]{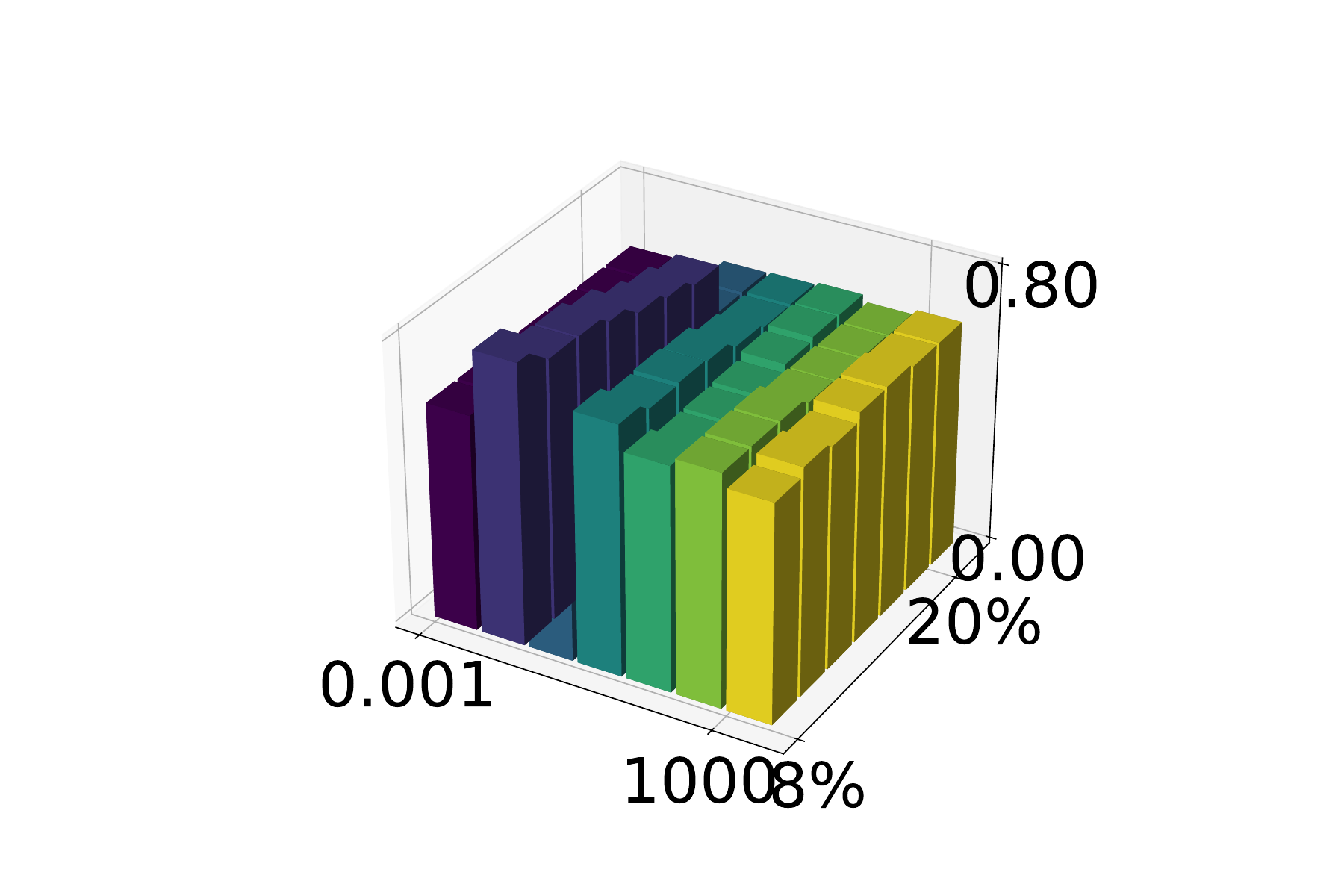}}
        \caption{$\alpha$}
        \label{psa}
    \end{subfigure}
    \centering
    \hspace{0.0001\linewidth}
    \begin{subfigure}{0.32\linewidth}
        \centering
        \scalebox{1.3}{\includegraphics[width=\linewidth]{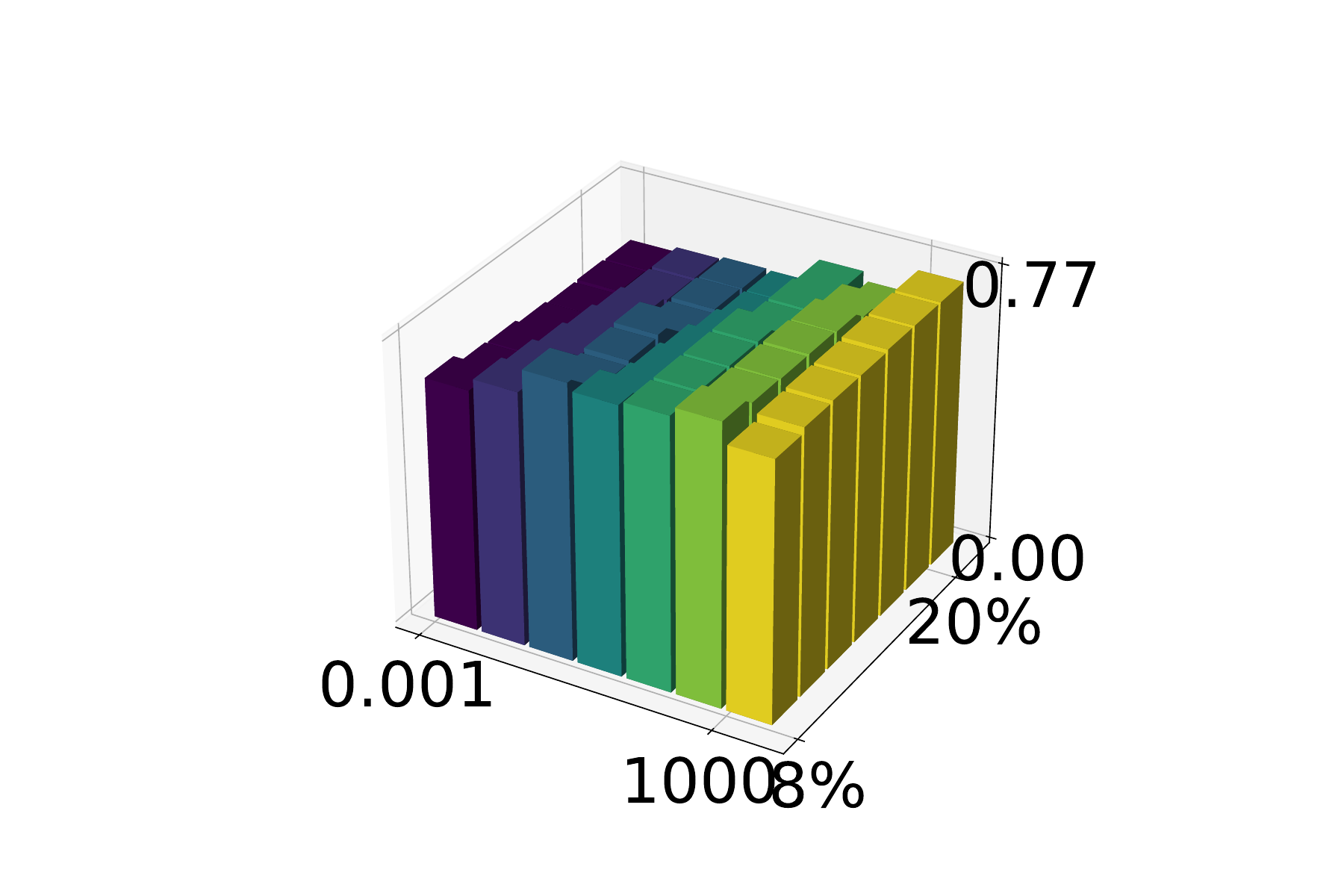}}
        \caption{$\beta$}
        \label{psb}
    \end{subfigure}
    \hspace{0.0001\linewidth}
    \begin{subfigure}{0.32\linewidth}
        \centering
        \scalebox{1.3}{\includegraphics[width=\linewidth]{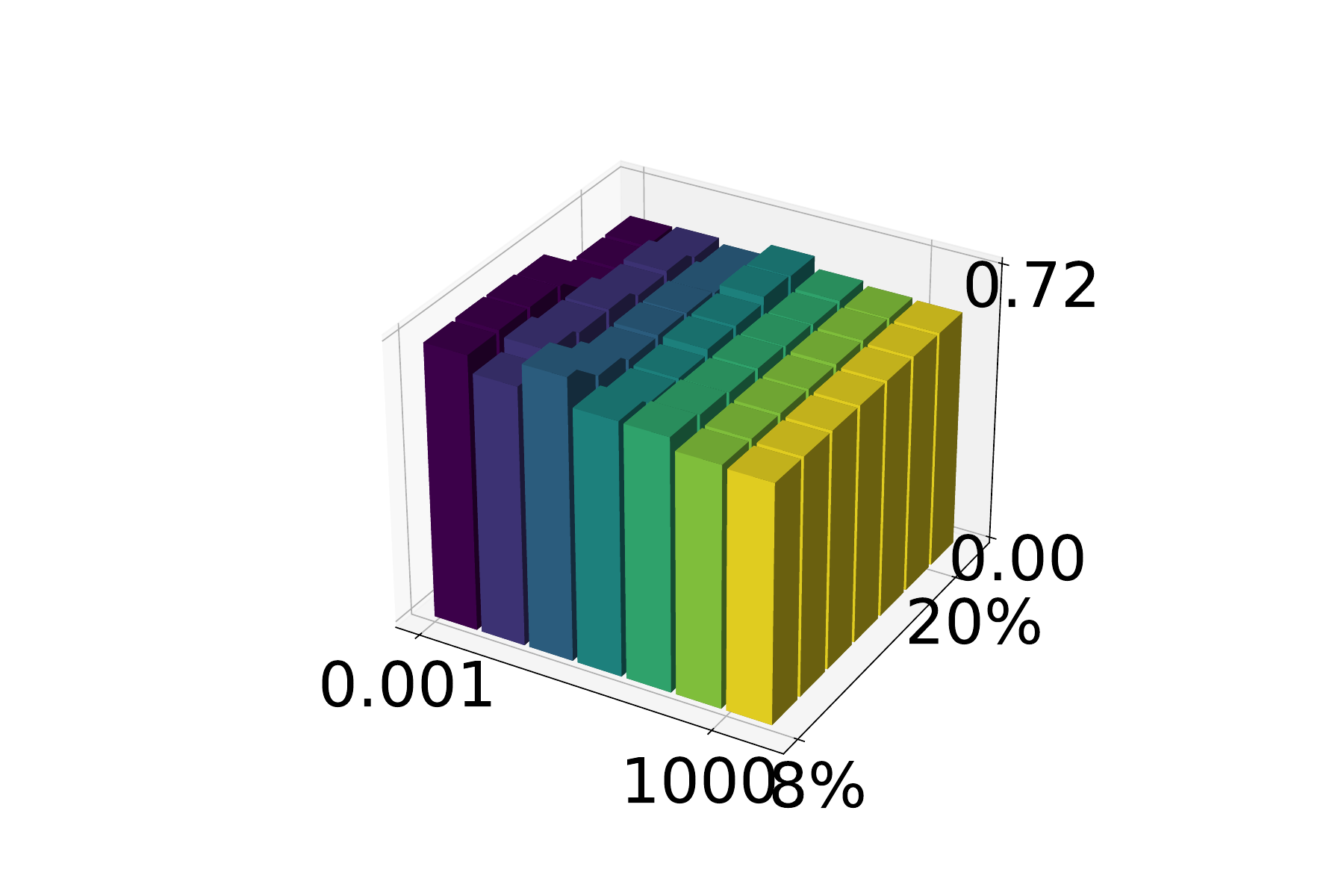}}
        \caption{$\gamma$}
        \label{psc}
    \end{subfigure}
    \caption{Parameter sensitivity studies on the \textbf{\textit{CAL}} in terms of Coverage.}
    \label{ps}
\end{figure}

\begin{figure}[htbp]
	\centering
	\begin{subfigure}{0.45\linewidth}
		\centering
		\includegraphics[width=\linewidth]{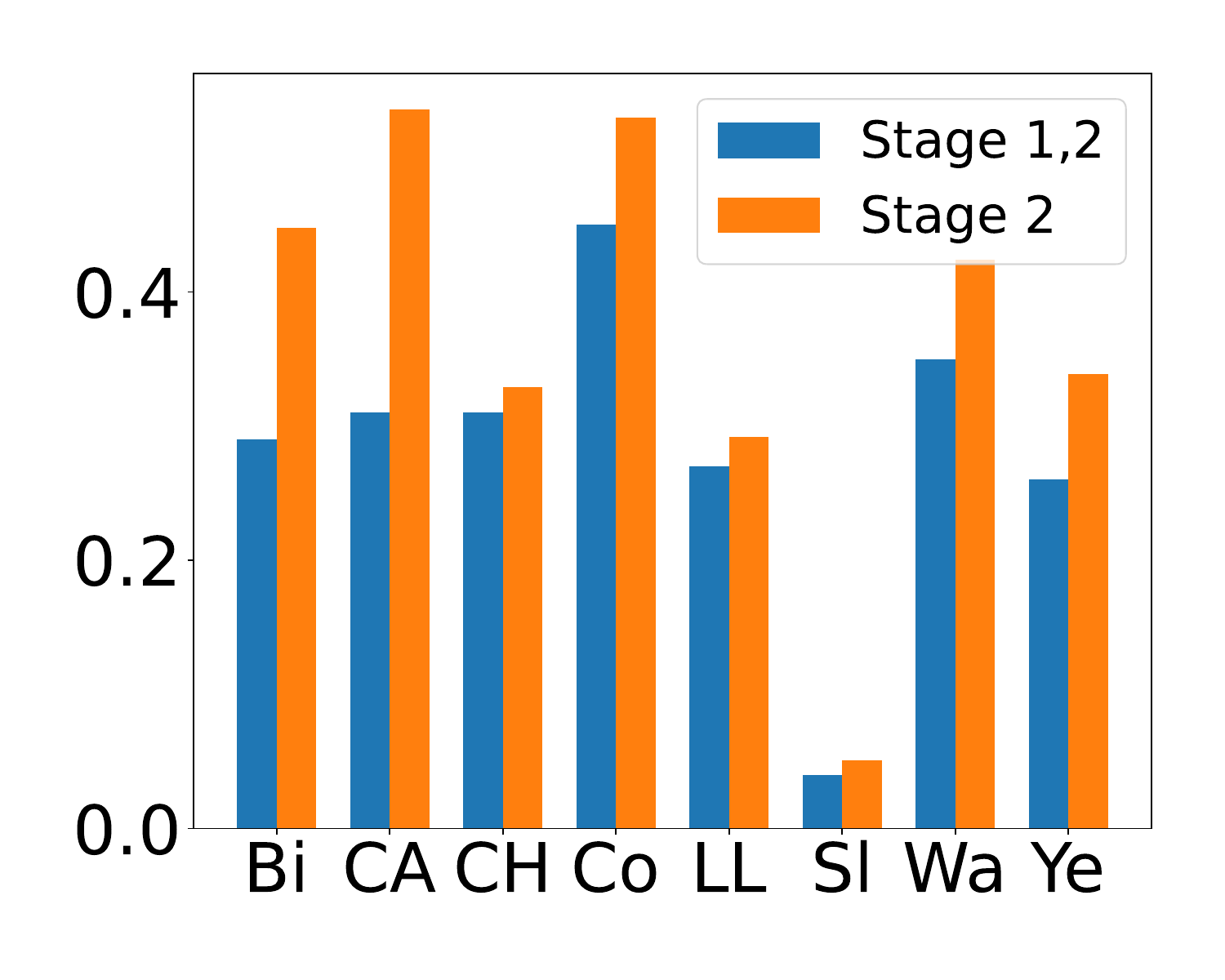}
		\caption{Ranking Loss}
		\label{3a}
	\end{subfigure}
	\centering
	\begin{subfigure}{0.45\linewidth}
		\centering
		\includegraphics[width=\linewidth]{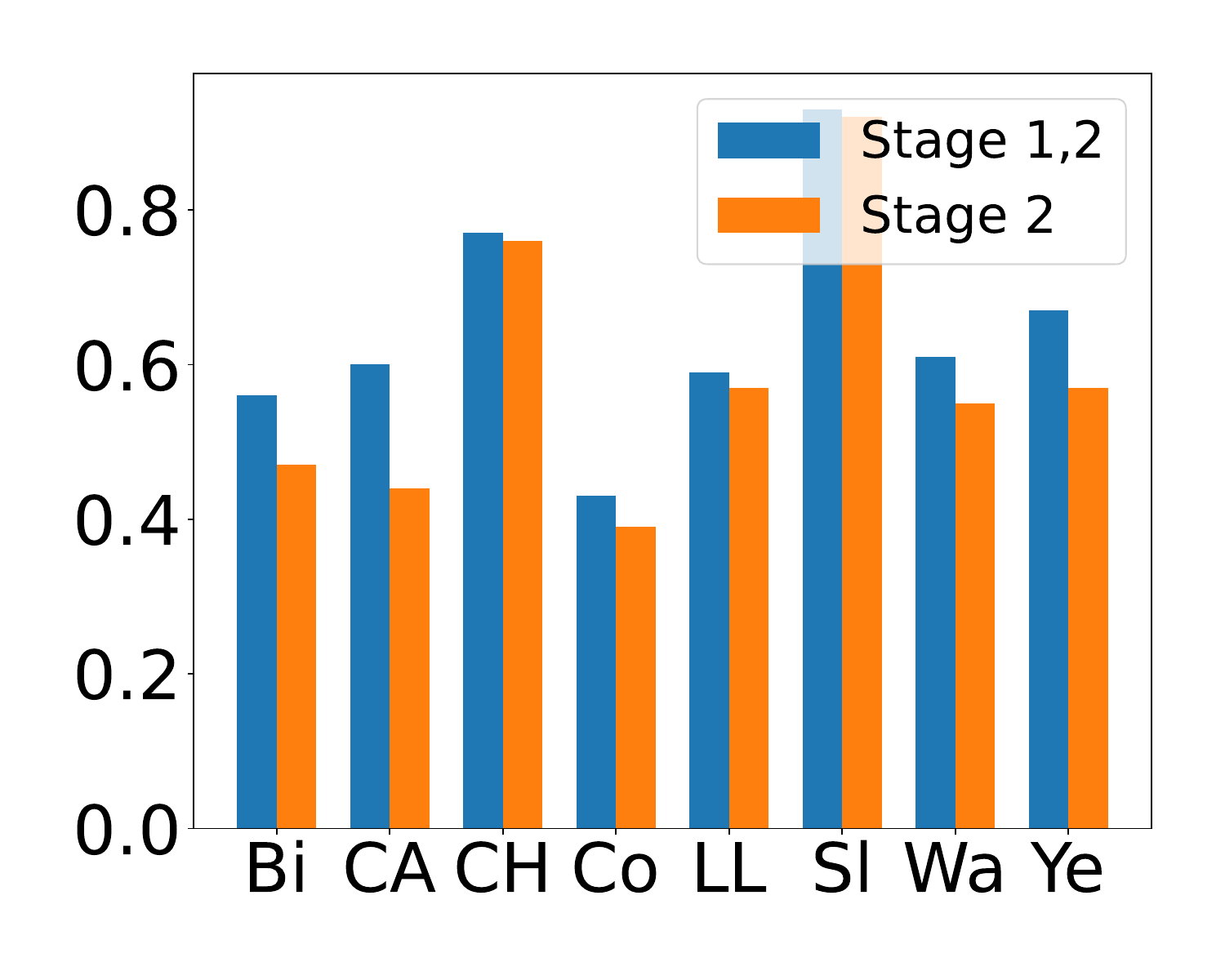}
		\caption{Average Precision}
		\label{3c}
	\end{subfigure}
	\caption{The results of the comparison experiment between the first two stage and the second stage.}
	\label{f3}
\end{figure}

\begin{figure}[htbp]
	\centering
	\begin{subfigure}{0.45\linewidth}
		\centering
		\includegraphics[width=\linewidth]{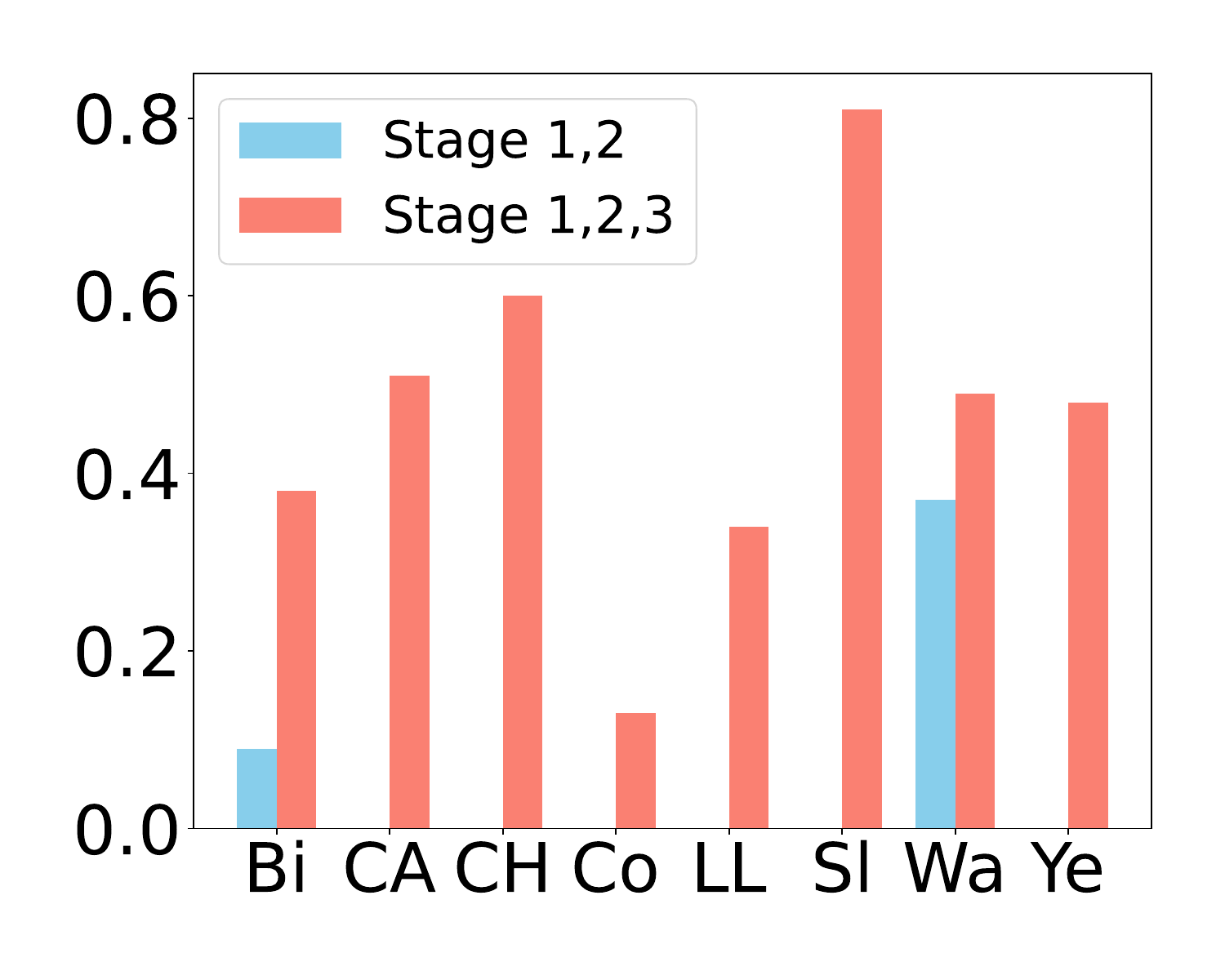}
		\caption{Mirco-F1}
		\label{4a}
	\end{subfigure}
	\centering
	\begin{subfigure}{0.45\linewidth}
		\centering
		\includegraphics[width=\linewidth]{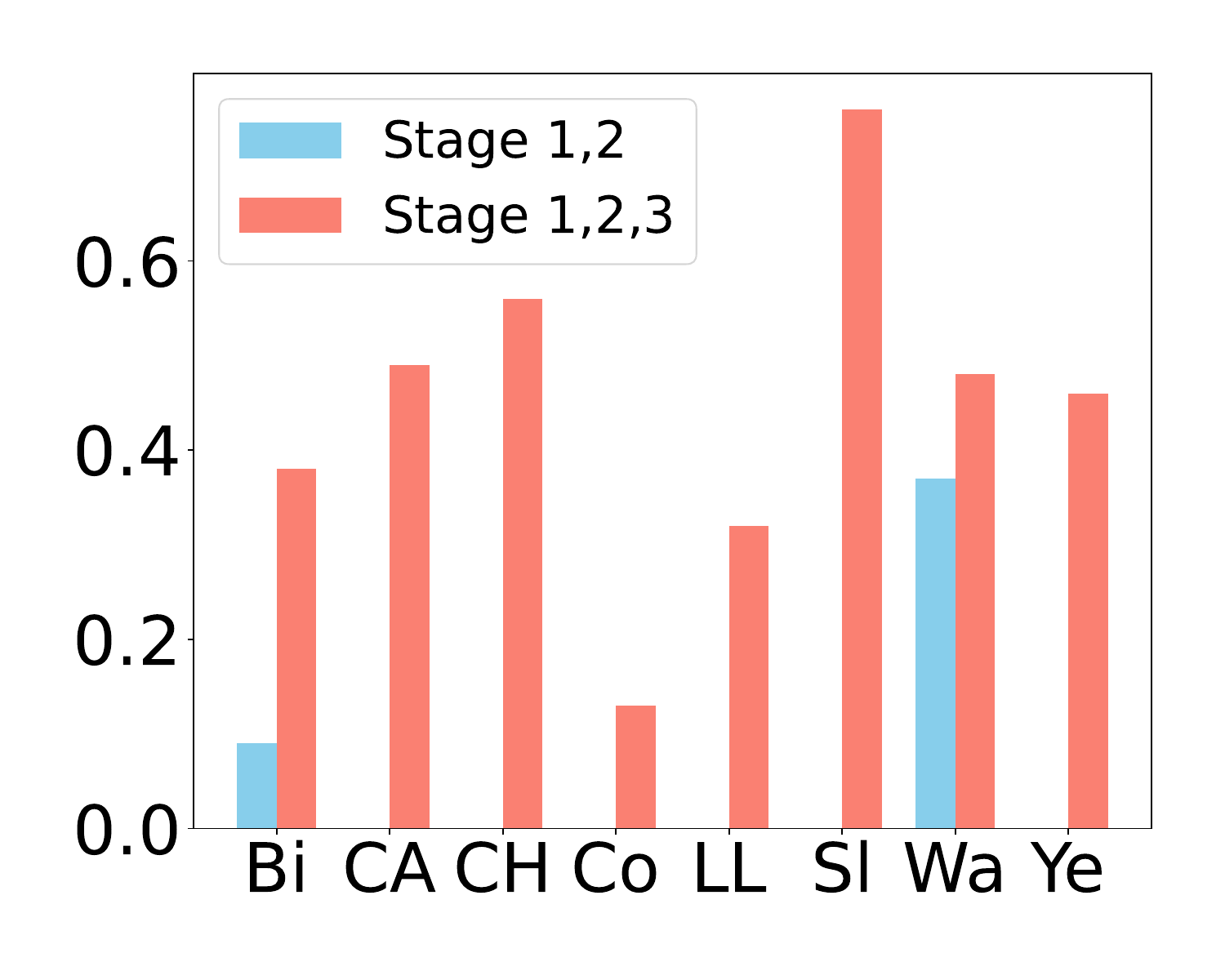}
		\caption{Marco-F1}
		\label{4b}
	\end{subfigure}
	
	\caption{The results of the comparison experiment between the first two stage and the whole process. }
	\label{f4}
\end{figure}

\subsection{Parameter Analysis}
In PML-FSMIR, there are three parameters $\alpha$, $\beta$, and $\gamma$ that affect  experimental results. Figure \ref{ps} shows how these three parameters affect the performance of the model on the \textbf{\textit{LLOG\_F}} in terms of Ranking Loss. Each parameter is independently tuned from 0.001 to 1000, and we selected the performance of the model when selecting 8\% of the features to 20\% of the features. From the figure, it can be seen that the model is obviously insensitive to the parameters, which shows the robustness of the model. 
\subsection{Ablation Study}
To prove the necessity of reconstructing objective matrix using  mutual information matrix in  first stage and  third stage, we set up two ablation experiments: (1) we employ the first two stage compared to directly conducting the second stage to verify the superiority of the first stage; (2) we employ whole procedure compared to conducting the first two stage to verify the superiority of the third stage. In the first experiments we employ Ranking Loss, Coverage, and Average Precision as evaluate metrics. In  the second experiments Mirco-F1 and Marco-F1 are used as evaluate criteria. The result are shown in Figures \ref{f3}, \ref{f4}. The Figure \ref{f3} proves the effectiveness of first stage. In all the datasets, the results of the first two stage are better than the second stage, which indicates that using mutual information matrix to reconstruct label matrix can effectively reduce the influence of noises in the labels.

The Figure \ref{f4} proves the effectiveness of the third stage. It clearly shows the significant improvement of weight matrix reconstruction. The method with the first two stage even cannot identify positive labels in six out of eight datasets while after  the third stage  it can do so in all datasets. This experiment enhances the correctness of our core idea: there are some labels in the label set that are more important than others. Finding features related to these labels can effectively improve the model's ability to identify positive labels.

\section{Conclusion}

In this paper, we tackle the challenges of label noise and sparsity in partial multi-label data by introducing a novel three-stage feature selection method PML-FSMIR). This method first reduce the noises in the label sets by reconstructing label matrix using mutual information matrix. Then it trains a weight matrix under a reformed low-rank assumption, which helps to prevent overfitting and ensures a more accurate reflection of the data’s underlying structure. Finally the weight matrix is reconstructed to enhance the ability of the selected features in effectively identifying key labels, thereby improving the model’s overall performance. Extensive experimental results validate the superiority of our method.  In the future, we plan to explore how to exploit mutual information and other methods to reduce noises and improve the identification of positive labels furthermore.

\section*{Acknowledgments}
This work was supported by the Science Foundation of Jilin Province of China under Grant YDZJ202501ZYTS286, and in part by Changchun Science and Technology Bureau Project under Grant 23YQ05.
\bigskip

\bibliographystyle{named}

\bibliography{ijcai25}

\end{document}